\documentclass[conference, a4paper]{IEEEtran}
\IEEEoverridecommandlockouts

\usepackage{cite}
\usepackage{amsmath,amssymb,amsfonts}
\usepackage{algorithmic}
\usepackage{graphicx}
\usepackage{textcomp}
\usepackage{xcolor}
\usepackage{subfigure}
\def\BibTeX{{\rm B\kern-.05em{\sc i\kern-.025em b}\kern-.08em
    T\kern-.1667em\lower.7ex\hbox{E}\kern-.125emX}}
\begin{document}

\title{Seed Selection for Human-Oriented Image Reconstruction via Guided Diffusion\\

}

\author{\IEEEauthorblockN{Yui Tatsumi}
\IEEEauthorblockA{\textit{Graduate School of FSE,} \\
\textit{Waseda University}\\
Tokyo, Japan \\
yui.t@fuji.waseda.jp}
\and
\IEEEauthorblockN{Ziyue Zeng}
\IEEEauthorblockA{\textit{Graduate School of FSE,} \\
\textit{Waseda University}\\
Tokyo, Japan \\
zengziyue@fuji.waseda.jp}
\and
\IEEEauthorblockN{Hiroshi Watanabe}
\IEEEauthorblockA{\textit{Graduate School of FSE,} \\
\textit{Waseda University}\\
Tokyo, Japan \\
hiroshi.watanabe@waseda.jp}
}

\maketitle

\begin{abstract}
Conventional methods for scalable image coding for humans and machines require the transmission of additional information to achieve scalability. A recent diffusion-based approach avoids this by generating human-oriented images from machine-oriented images without extra bitrate. However, it utilizes a single random seed, which may lead to suboptimal image quality. In this paper, we propose a seed selection method that identifies the optimal seed from multiple candidates to improve image quality without increasing the bitrate. To reduce the computational cost, selection is performed based on intermediate outputs obtained from early steps of the reverse diffusion process. Experimental results demonstrate that our proposed method outperforms the baseline, which uses a single random seed without selection, across multiple evaluation metrics.

\end{abstract}


\section{Introduction}
The rapid advancement in deep learning has led to a growing number of scenarios where images are usually analyzed by recognition models and occasionally checked by humans, such as traffic surveillance. However, these two objectives require distinct types of images and therefore codecs need to be optimized for each. Learned Image Compression (LIC) for human viewing focuses on preserving perceptual details. In contrast, Image Coding for Machines (ICM) discards information irrelevant to recognition tasks. To bridge this gap, scalable image compression techniques that simultaneously support both human visual perception and machine vision have been studied. 

Some existing approaches to scalable image coding have addressed this challenge by transmitting supplementary information. In this method, machine-oriented image is first decoded. When human viewing is required, additional information that allows the machine-oriented image to be converted into a human-oriented one is transmitted. While effective, these methods inevitably increase the overall bitrate.

A novel diffusion-based approach offers an alternative by converting machine-oriented images into human-oriented ones without transmitting any additional data with a guided diffusion model. This method demonstrates superior performance in perceptual metrics. However, although diffusion models are known to produce highly diverse outputs depending on the random seed, the method employs only a single random seed, which may lead to suboptimal image quality.

In this paper, we propose a method that improves the quality of generated images by selecting the optimal seed through multiple reverse diffusion processes. To reduce computation, we further propose a strategy that selects the optimal seed based on intermediate outputs from early reverse diffusion steps. 

\begin{figure}[t]
  \centering
\includegraphics[width=\hsize]
{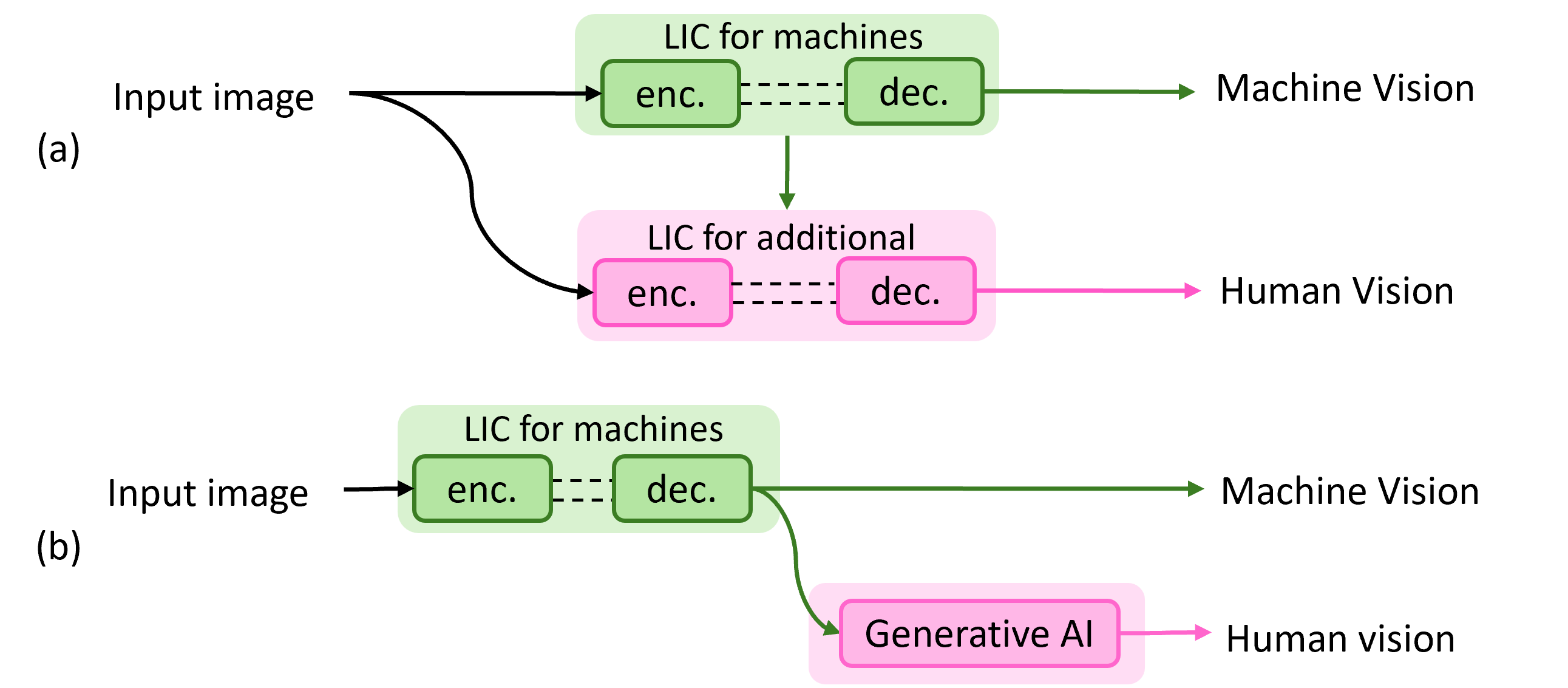}
  \caption{Overview of scalable image coding pipelines. (a) LIC-based conventional methods, (b) Diffusion-based methods.}
  \label{fig:conventional}
\end{figure}

\begin{figure*}[t]
  \centering
\includegraphics[width=0.7\hsize]
{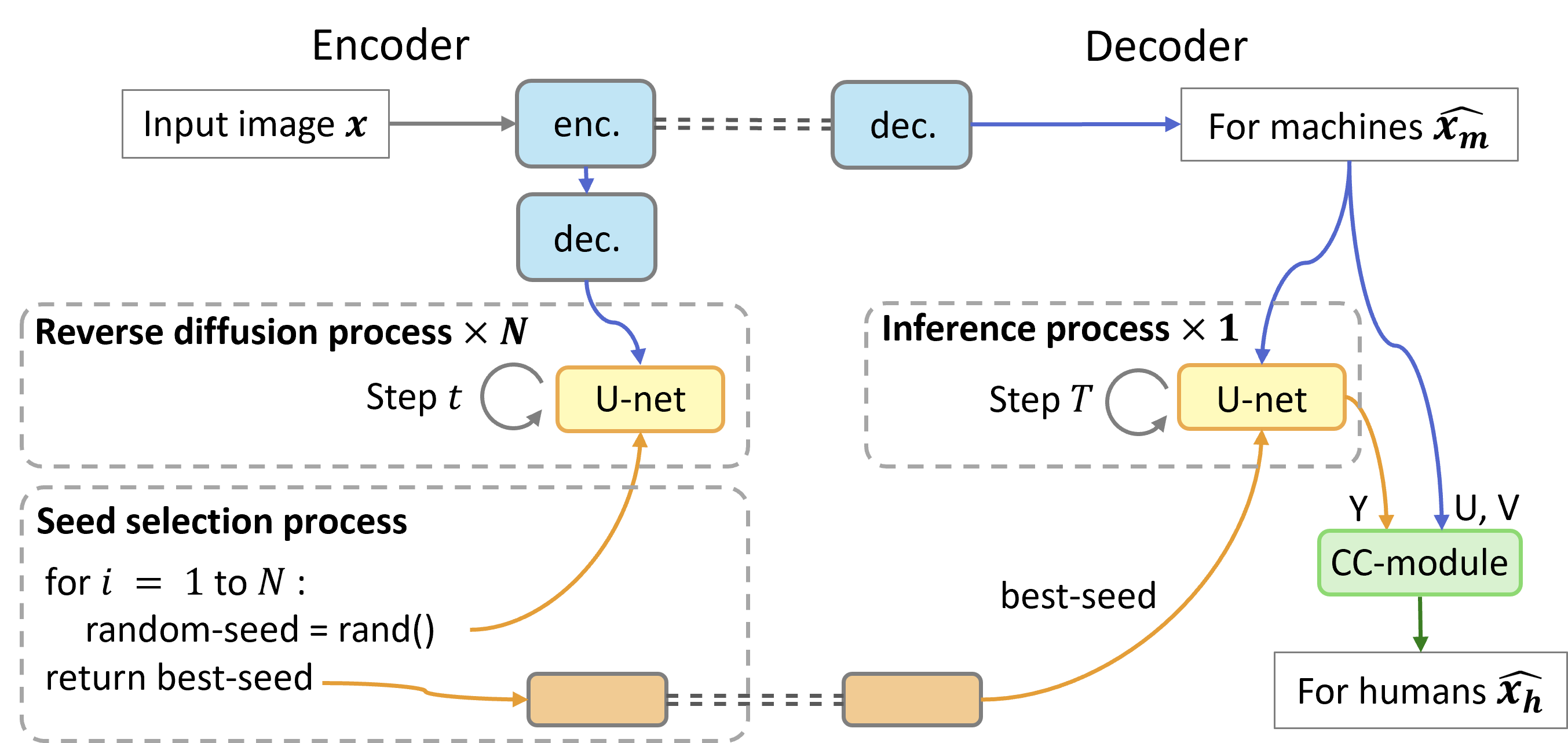}
  \caption{Processing flow of the proposed seed selection method.}
  \label{fig:overview}
\end{figure*}

\section{Related Work}
\subsection{Scalable Image Coding for Humans and Machines}
Research on scalable image coding for humans and machines has been explored. ICMH-FF \cite{ICMH-FF} combines two distinct LIC models: one optimized for machine vision, and another that decodes additional information for human perception. These two models are integrated by a feature fusion network, which enables the effective combination of features. Building upon this concept, FR-ICMH and PR-ICMH \cite{FR and PR} further improve compression performance by explicitly utilizing feature-level and pixel-level residuals between machine-oriented and human-oriented images, respectively. The processing flow of these conventional scalable codecs is shown in Fig. \ref{fig:conventional}(a). While these methods demonstrate high compression performance in objective quality metrics such as PSNR, they require the transmission of a substantial amount of additional data, which increases the total bitrate. 

\subsection{Diffusion-based Image Coding}
Diffusion models have recently emerged as powerful generative tools capable of synthesizing high-quality images from minimal inputs, such as text prompts. In the context of image coding, H. Watanabe \textit{et al.} \cite{pcsj} propose a prompt-based image coding scheme that generates images using text prompts and edge information by a diffusion model. In this method, the quality of the generated image is maximized by applying multiple parameters such as random seeds and prompts.

Furthermore, a diffusion-based approach to scalable image coding \cite{guided-diffusion} has been proposed, which converts machine-oriented images into human-oriented images without transmitting any additional bitrate. The overall pipeline is illustrated in Fig. \ref{fig:conventional}(b). This method leverages ICM-decoded images as conditional inputs and uses Stable Diffusion \cite{stable diffusion} with ControlNet \cite{controlnet} to generate human-oriented images. It also introduces a Color Controller (CC) module to improve color fidelity by replacing the color components of the generated image with those from the ICM-decoded image, while preserving luminance.
Although this method demonstrates high performance especially in perceptual quality metrics compared to traditional scalable codecs, the control parameters, such as random seed, have not been thoroughly considered, and the reconstructed image may therefore be further optimized.

\subsection{Impact of Random Seed in Diffusion Models}
In diffusion-based generative models, the random seed plays a critical role by controlling the initial noise and intermediate noise steps. K. Xu \textit{et al.} \cite{good seed} investigate the impact of random seeds in text-to-image diffusion models. Their analysis reveals that the choice of seeds significantly influences image quality, style, and composition, and affects the score of perceptual quality metrics. They further demonstrate that images generated from different seeds are highly distinguishable, and that certain seeds consistently lead to specific visual patterns. These findings emphasize the practical importance of seed selection in diffusion-based image generation.

\begin{figure*}[t]
  \centering
\includegraphics[width=\hsize]
{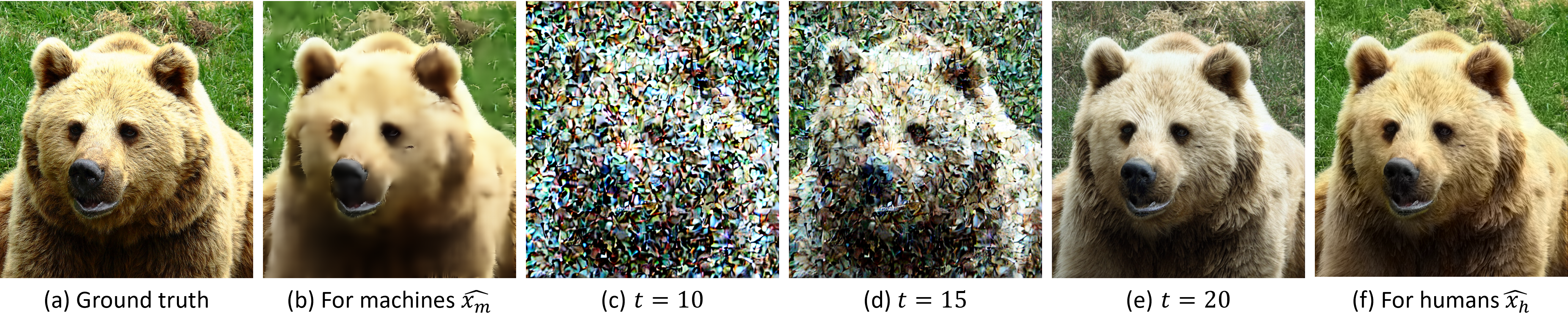}
  \caption{Examples of original and decoded images: (a) Original image, (b) Decoded image for machines using SA-ICM, (c) Output image at $t=10$, (d) Output image at $t=15$, (e) Output image at $t=20$, (f) Final reconstructed image for humans after applying the CC-module.}
  \label{fig:output}
\end{figure*}

\section{Proposed Method}
\subsection{Overview}
We propose an early-step seed selection method to improve the quality of human-oriented images generated from machine-oriented ones using guided diffusion. As diffusion models are highly sensitive to the choice of random seed, selecting an optimal seed is critical for reconstructing the image with high fidelity to the original one. However, evaluating multiple seeds through full diffusion processes is computationally expensive. To address this, we introduce a strategy that identifies the optimal seed based on intermediate outputs obtained at early steps of the reverse diffusion process. An overview of the proposed framework is illustrated in Fig. \ref{fig:overview}. The process can be divided into two stages: seed selection at the encoder side and full reconstruction at the decoder side.

\subsection{Seed Selection Process}
The original image is first input into an ICM model to obtain an image for machines. This machine-oriented image is then utilized as a condition to generate $N$ candidate outputs using a ControlNet-conditioned Stable Diffusion model, each with a different random seed. Instead of completing the entire reverse diffusion process, generation is performed only up to an intermediate timestep $t$.
The similarity of each candidate image to the ground truth is evaluated and the seed corresponding to the most similar output is selected. Unlike typical image generation tasks, our setting allows for objective seed selection since the encoder has access to the original image to compare against.

\subsection{Image Reconstruction Process}
The selected seed is then transmitted to the decoder. Since the size of the seed is negligible compared to that of additional data in conventional scalable codecs, the resulting bitrate is virtually zero. At the decoder side, full reverse diffusion with $T$ steps is executed only one time using the received seed to generate the human-oriented image. Finally, a CC-module is applied to enhance color fidelity by injecting the chrominance components from the ICM-decoded image.

\section{Experiment}
\subsection{Experimental Details}
We evaluate our proposed method on COCO dataset \cite{coco}. SA-ICM \cite{SA-ICM}, which was pretrained on 118,287 COCO-train images, is utilized as the ICM model. An example of a machine-oriented image decoded by SA-ICM is shown in Fig. \ref{fig:output}(b). As illustrated, fine textures and other details necessary for human viewing are significantly reduced. These SA-ICM coded COCO-train images are used as conditional inputs to a ControlNet and trained to reproduce the original image. The text prompt is kept as an empty string throughout the experiments.
During the experiment, we use 5,000 images from COCO-val dataset. Each image is first coded by SA-ICM and input into the pretrained ControlNet. For each image, five candidate images are generated with different random seeds. The total number of reverse diffusion steps is set to $T=20$. To reduce computational cost, seed selection is performed at timesteps $t ={1, 5, 10, 15, 18, 20}$, based on the Y-channel PSNR between each candidate and the original image. The seed corresponding to the highest PSNR is then selected and used for the full inference and a CC-module is employed. Finally, the quality of the generated images is evaluated with PSNR, SSIM, and LPIPS \cite{lpips}. 
As a baseline, we average the evaluation scores of 20 images generated with different random seeds, simulating the generation with a single random seed without seed selection.
The results are also compared with LIC-TCM \cite{LIC-TCM}, a LIC model optimized for human viewing, as well as with conventional scalable coding methods: ICMH-FF, FR-ICMH, and PR-ICMH. 

\subsection{Experimental Results}
Example outputs at various reverse diffusion steps are shown in Fig. \ref{fig:output}(c) - \ref{fig:output}(e). As the step progresses, textures and edges are refined, and noise is gradually removed. Finally, Fig. \ref{fig:output}(f) shows the human-oriented image reconstructed using the selected seed and enhanced with the CC-module. 
Table \ref{tab:results} summarizes the quantitative performance of seed selection at different reverse diffusion steps. As shown, our method consistently improves over the random baseline across all metrics. Notably, PSNR steadily improves as $t$ increases, indicating that later reverse diffusion steps provide more reliable guidance for seed selection. In contrast, SSIM and LPIPS remain relatively stable across timesteps. 
The rightmost column represents the average time per input image required to generate candidate outputs for seed selection at the encoder side, when the reverse diffusion process is performed up to each timestep $t$. Since the baseline using a random seed does not involve seed selection, the corresponding entry is left blank. The generation time increases approximately linearly with the number of reverse diffusion steps. 
For instance, seed selection at $t = 10$ requires approximately half the computation time of full-step selection at $t = 20$, while producing slightly lower image quality.
This demonstrates a trade-off between computational efficiency and reconstruction performance.

\begin{table}[t]
    \begin{center}
    \caption{Comparison of seed selection strategies at different timesteps}
    \label{tab:results}
    \begin{tabular}{c|cccc}
        \hline
        Selection Timestep ($t$) & PSNR(dB)$\uparrow$ & SSIM$\uparrow$ & LPIPS$\downarrow$ & Time(s)\\ \hline 
        \hline
        Random (no selection)  & 21.72 & 0.554 & 0.225 & -\\
        $1$  & 22.04 & 0.565 & 0.220 & 0.295\\
        $5$  & 22.13 & 0.569 & 0.217 & 1.735\\
        $10$  & 22.21 & \underline{0.574} & 0.215 & 3.425\\
        $15$  & 22.28 & \textbf{0.576} & \textbf{0.213} & 5.270\\
        $18$  & \underline{22.41} & \textbf{0.576} & \underline{0.214} & 6.450\\
        $20$ (full) & \textbf{22.50} & 0.572 & 0.217 & 7.381\\
        \hline
    \end{tabular}
    \end{center}
\end{table}

\begin{figure*}
\centering
\subfigure[PSNR (↑)]{%
\includegraphics[clip, width=0.31\hsize]{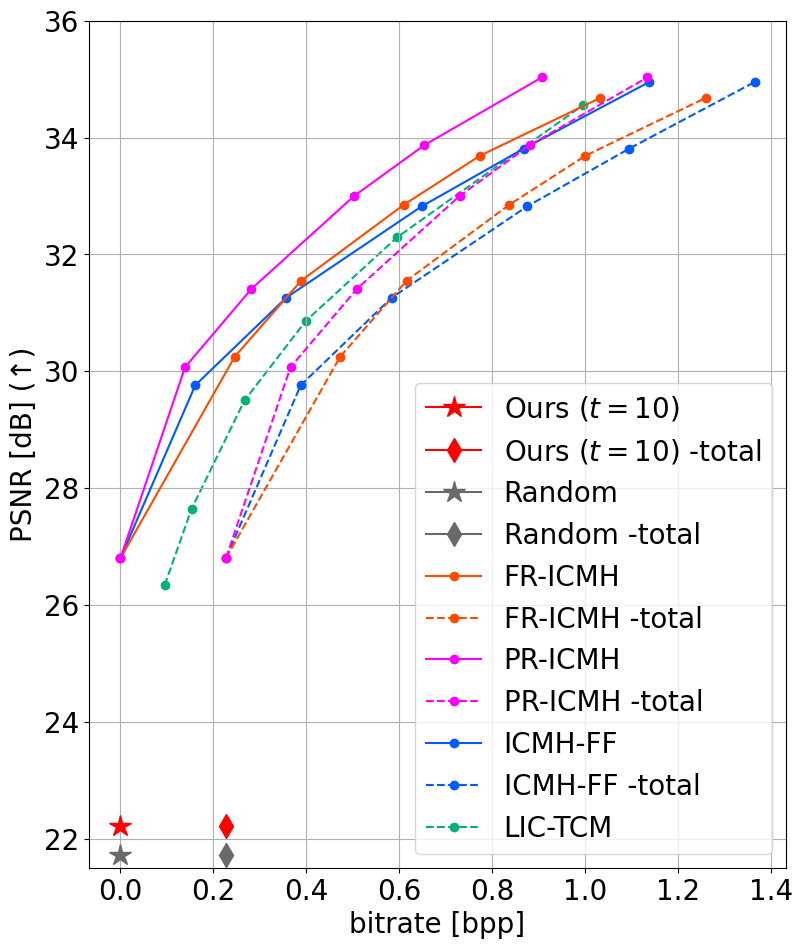}}%
\hspace{0.0025pt}
\subfigure[SSIM (↑)]{%
\includegraphics[clip, width=0.32\hsize]{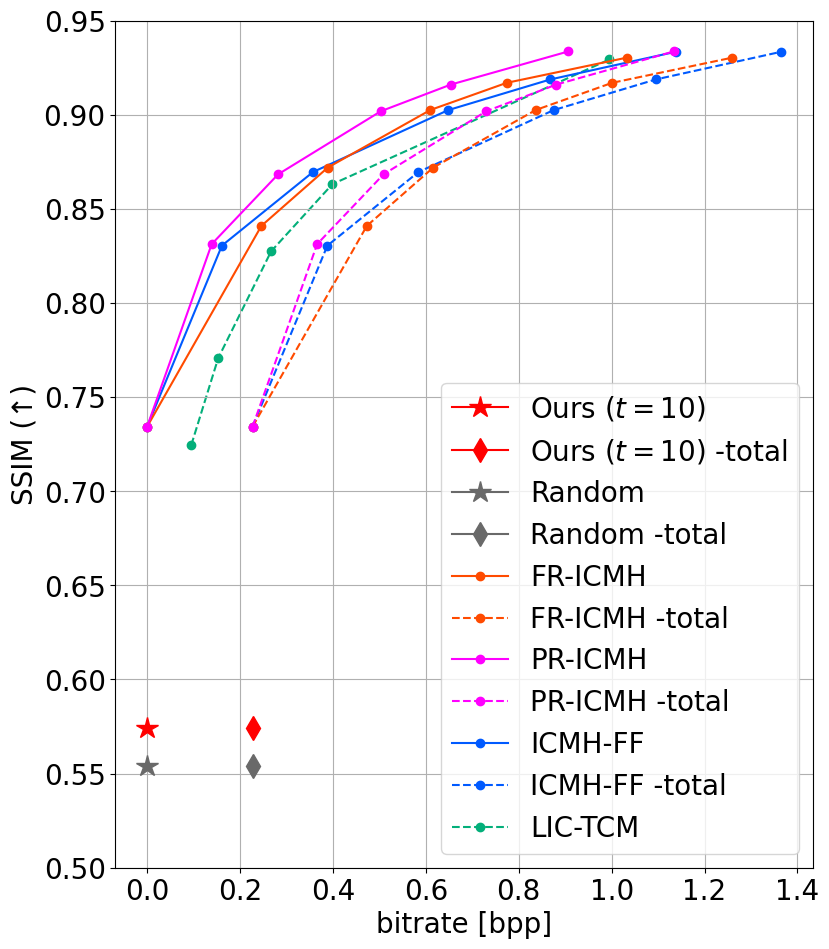}}%
\hspace{0.0025pt}
\subfigure[LPIPS (↓)]{%
\includegraphics[clip, width=0.32\hsize]{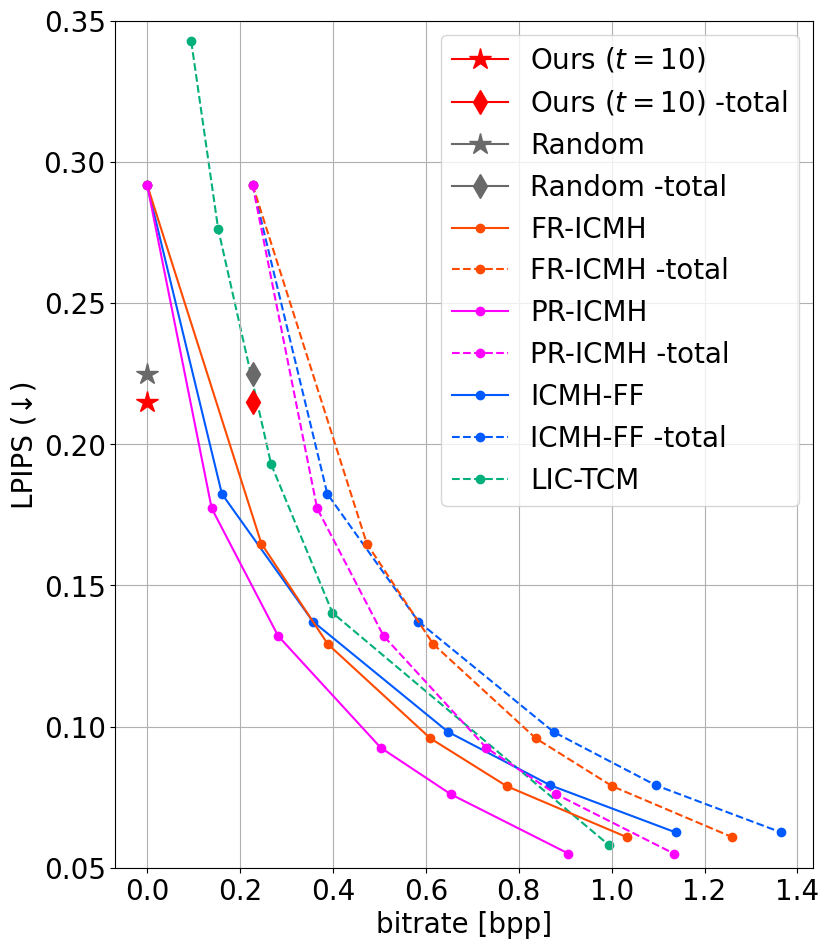}}%
\caption{Compression performance for humans of proposed and comparative methods evaluated using (a) PSNR, (b) SSIM, and (c) LPIPS. }
\label{fig:rd-curve}
\end{figure*}

\begin{figure}[t]
  \centering
\includegraphics[width=0.95\hsize]
{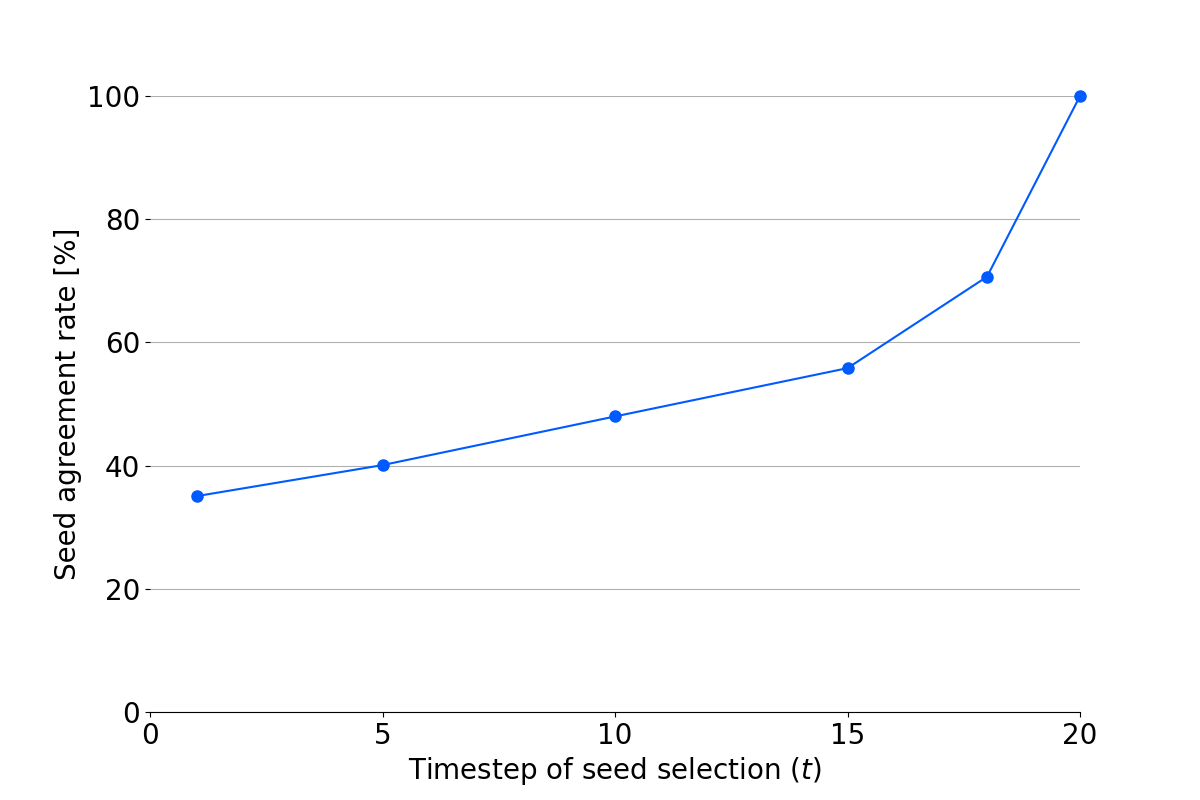}
  \caption{Agreement rate between early-step selected seeds and the optimal seed at $t=20$.}
  \label{fig:match}
\end{figure}

Fig. \ref{fig:rd-curve} illustrates the compression performance for human perception across various methods, evaluated using three metrics. Solid lines and star-shaped markers indicate the bitrate required for transmitting additional information for humans only, while dashed lines and diamond-shaped markers represent the total bitrate, including the information necessary for decoding images for both machines and humans. In this figure, the red markers correspond to our proposed method with seed selection performed at an early timestep $t = 10$, and the gray markers represent the baseline using a single random seed without selection.
Fig. \ref{fig:rd-curve}(a) and \ref{fig:rd-curve}(b) reveal that our proposed method is less effective than conventional methods in metrics that measure the pixel-level fidelity. In contrast, Fig. \ref{fig:rd-curve}(c) shows that our method demonstrates better perceptual similarity when additional bitrate is 0 [bpp].

To further evaluate the effectiveness of early-step seed selection, we measure its agreement with the optimal seed chosen after full inference, $t = 20$. Fig. \ref{fig:match} illustrates the agreement rate between the selected seed at each timestep and the final optimal seed. Notably, even at $t=1$, the agreement rate exceeds 20\%, which is the expected value under random selection from five candidates. The results indicate that outputs at early-steps already contain informative cues although it is still in the denoising phase. Moreover, the agreement rate increases approximately linearly until $t = 15$, and then rises more sharply. This result indicates that later diffusion steps provide more reliable signals for seed selection.

\section{Conclusion}
In this paper, we propose a seed selection method for diffusion-based reconstruction of human-oriented images from machine-oriented inputs while maintaining a zero-bitrate overhead. By evaluating early-step outputs, our method selects the optimal seed with reduced computational cost. Experimental results show that our proposed method improves image quality over the baseline using a single random seed. In practical applications, the observed trade-off between reconstruction quality and computation time should be considered.
Future work includes exploring adaptive timestep strategies for more efficient seed selection.

\section*{Acknowledgment}
The results of this research were obtained from the commissioned research (JPJ012368C05101) by National Institute of Information and Communications Technology (NICT), Japan.

\end{document}